\documentclass{article}

\usepackage[table,xcdraw]{xcolor}

    \usepackage[preprint]{neurips_2025}

\usepackage{amsmath,amsfonts,bm}

\def\eqref#1{equation~\ref{#1}}

\def\1{\bm{1}}

\DeclareMathAlphabet{\mathsfit}{\encodingdefault}{\sfdefault}{m}{sl}
\SetMathAlphabet{\mathsfit}{bold}{\encodingdefault}{\sfdefault}{bx}{n}

\newcommand{\R}{\mathbb{R}}

\newcommand{\KL}{D_{\mathrm{KL}}}

\DeclareMathOperator*{\argmin}{arg\,min}

\newcommand{\cut}[1]{}

\usepackage[framemethod=TikZ]{mdframed}
\mdfdefinestyle{MyFrame}{%
    linecolor=black,
    outerlinewidth=0.5pt,
    roundcorner=1pt,
    innertopmargin=2pt, 
    innerbottommargin=2pt, 
    innerrightmargin=7pt,
    innerleftmargin=7pt,
    backgroundcolor=black!0!white}

\mdfdefinestyle{MyFrame2}{%
    linecolor=white,
    outerlinewidth=1pt,
    roundcorner=2pt,
    innertopmargin=10pt,
    innerbottommargin=0.5pt,
    innerrightmargin=10pt,
    innerleftmargin=10pt,
    backgroundcolor=black!3!white}

\mdfdefinestyle{MyFrameEq}{%
    linecolor=white,
    outerlinewidth=0pt,
    roundcorner=0pt,
    innertopmargin=0pt,
    innerbottommargin=0pt,
    innerrightmargin=7pt,
    innerleftmargin=7pt,
    backgroundcolor=black!3!white}

\mdfdefinestyle{FrameAdded}{%
    linecolor=black,
    outerlinewidth=0pt,
    roundcorner=0pt,
    innertopmargin=0pt,
    innerbottommargin=0pt,
    innerrightmargin=0pt,
    innerleftmargin=0pt,
    backgroundcolor=PastelGreen}

\newcommand{\AND}{\texttt{{\color{pastel_purple} \textbf{AND}}}\xspace}

\definecolor{customwhite}{HTML}{FCFBF7}
\definecolor{customturq}{HTML}{1D9D79}
\definecolor{customorange}{HTML}{D96002} 
\definecolor{custombeige}{HTML}{d6c9b1} 

\definecolor{custompurple}{HTML}{AEADF0}
\definecolor{customwhite2}{HTML}{fbf9f4}
\definecolor{customblue}{HTML}{4D9DE0} 

\usepackage{silence}
\usepackage[utf8]{inputenc}
\usepackage[T1]{fontenc}
\usepackage{tabularx}
\usepackage{hyperref}
\usepackage{lipsum}
\usepackage{url}
\usepackage{placeins}
\usepackage{booktabs}
\usepackage{soul}
\usepackage{amsfonts}
\usepackage{nicefrac}
\usepackage{microtype}
\usepackage{natbib}
\usepackage{proba}
\usepackage{wrapfig}
\usepackage{caption}
\usepackage{amssymb}
\usepackage{pifont}

\usepackage{subcaption}
\usepackage{float}
\usepackage{rotating}
\usepackage{etoolbox}
\usepackage{xparse}
\usepackage{grffile}
\usepackage{amsmath}
\usepackage{xspace}
\usepackage{euscript}
\usepackage{enumitem}
\usepackage{mathtools}
\usepackage{tikz}
\usepackage{tablefootnote}
\usepackage[flushleft]{threeparttable}
\usepackage{multirow}
\usepackage[title]{appendix}
\usepackage{arydshln}
\usepackage{array, booktabs}
\usepackage{comment}
\usepackage{graphicx}
\usepackage{adjustbox}
\usepackage{dsfont}
\usepackage{amsmath,amsthm}
\usepackage{balance}
\usepackage{multicol}
\usepackage{nameref}
\usepackage{pdfpages}
\usepackage{braket}
\usepackage[normalem]{ulem}
\usepackage{bbding}
\usepackage[capitalise]{cleveref}   
\usepackage{xfrac}
\usepackage[usestackEOL]{stackengine}
\usepackage{indentfirst}
\usepackage{csquotes}
\usepackage{pgf}
\usepackage{varwidth}
\usepackage{verbatimbox}
\usepackage{verbatim}
\usepackage{bm}
\usepackage{algorithmic}
\usepackage[colorinlistoftodos,prependcaption,textsize=small]{todonotes}

\usepackage{empheq}
\definecolor{myblue}{rgb}{.8, .8, 1}

\definecolor{pastelblue}{RGB}{76,113,175}
\definecolor{pastelgreen}{RGB}{144,238,144}
\definecolor{pastelred}{RGB}{196,78,82}
\definecolor{pastelgrey}{RGB}{230,230,230}
\definecolor{pastelbeige}{RGB}{243,236,221}
\definecolor{pastelpurple}{RGB}{154,139,192}
\definecolor{salmon}{RGB}{250, 128, 114}
\definecolor{darkgreen}{rgb}{0,0.6,0}
\definecolor{darkred}{rgb}{0.5,0,0}
\definecolor{verylightgreen}{HTML}{F6FFF9}
\definecolor{verylightred}{HTML}{FFF4F3}
\definecolor{verylightgray}{HTML}{F4F6F6}
\definecolor{babyblueeyes}{rgb}{0.63, 0.79, 0.95}
\definecolor{lightpink}{rgb}{1.00, 0.714, 0.757}

\usepackage{tikzpeople}
\usetikzlibrary{shapes,decorations,arrows,calc,arrows.meta,fit,positioning}

\tikzset{
    -Latex,auto,node distance =1 cm and 1 cm,semithick,
    state/.style ={ellipse, draw, minimum width = 0.7 cm},
    point/.style = {circle, draw, inner sep=0.04cm,fill,node contents={}},
    bidirected/.style={Latex-Latex,dashed},
    el/.style = {inner sep=2pt, align=left, sloped}
}

\makeatletter
\def\thmt@refnamewithcomma #1#2#3,#4,#5\@nil{%
	\@xa\def\csname\thmt@envname #1utorefname\endcsname{#3}%
	\ifcsname #2refname\endcsname
	\csname #2refname\expandafter\endcsname\expandafter{\thmt@envname}{#3}{#4}%
	\fi}
\makeatother
\Crefname{conjecture}{Conjecture}{Conjectures}
\Crefname{definition}{Definition}{Definitions}
\Crefname{observation}{Observation}{Observations}
\Crefname{assumption}{Assumption}{Assumptions}
\Crefname{axiom}{Axiom}{Axioms}
\Crefname{case}{Case}{Cases}
\Crefname{claim}{Claim}{Claims}
\Crefname{conclusion}{Conclusion}{Conclusions}
\Crefname{condition}{Condition}{Conditions}
\Crefname{criterion}{Criterion}{Criteria}
\Crefname{exercise}{Exercise}{Exercises}
\Crefname{example}{Example}{Examples}
\Crefname{notation}{Notation}{Notations}
\Crefname{problem}{Problem}{Problems}
\Crefname{property}{Property}{Properties}
\Crefname{remark}{Remark}{Remarks}
\Crefname{solution}{Solution}{Solutions}
\Crefname{summary}{Summary}{Summaries}
\Crefname{motivation}{Motivation}{Motivations}
\Crefname{query}{Query}{Queries}
\crefname{algocf}{Alg.}{Algs.}
\Crefname{algocf}{Algorithm}{Algorithms}

\newcommand{\one}{\mathds{1}}

\newcommand*\dbar[1]{\overline{\overline{\lower0.2ex\hbox{$#1$}}}}

\def\cB{{\mathcal{B}}}

\def\cD{{\mathcal{D}}}

\DeclareFontFamily{U}{BOONDOX-calo}{\skewchar\font=45 }
\DeclareFontShape{U}{BOONDOX-calo}{m}{n}{
  <-> s*[1.05] BOONDOX-r-calo}{}
\DeclareFontShape{U}{BOONDOX-calo}{b}{n}{
  <-> s*[1.05] BOONDOX-b-calo}{}
\DeclareMathAlphabet{\mathcalb}{U}{BOONDOX-calo}{m}{n}
\SetMathAlphabet{\mathcalb}{bold}{U}{BOONDOX-calo}{b}{n}
\DeclareMathAlphabet{\mathbcalb}{U}{BOONDOX-calo}{b}{n}

\usepackage{cancel}
\renewcommand{\paragraph}[1]{{\noindent \textbf{#1.}}}

\newcommand{\update}[0]{{\tt Update} \ }
\newcommand{\sample}[0]{{\tt Sample} \ }

\definecolor{pastel_purple}{HTML}{756FB3}
\definecolor{pastel_green}{HTML}{1D9D79}
\newcommand{\sampling}{%
    \texttt{{\color{pastel_green}\textbf{sampling}}}
}
\newcommand{\learning}{%
    \texttt{{\color{pastel_purple}\textbf{energy minimization}}}
}

\colorlet{PastelPurpleLight}{pastel_purple!15!white}
\colorlet{PastelGreenLight}{pastel_green!15!white}
\sethlcolor{PastelPurpleLight}
\sethlcolor{PastelGreenLight}
\newcommand{\colorcell}{\cellcolor{PastelGreenLight}}

\let\originalleft\left
\let\originalright\right
\renewcommand{\left}{\mathopen{}\mathclose\bgroup\originalleft}
\renewcommand{\right}{\aftergroup\egroup\originalright}

\global\long\def\norm#1{\left\lVert #1\right\rVert }
\global\long\def\inner#1#2{\left\langle #1, #2\right\rangle}

\definecolor{antiquefuchsia}{rgb}{0.57, 0.36, 0.51}
\definecolor{amethyst}{rgb}{0.6, 0.4, 0.8}

\newcommand{\deriv}[2]{\frac{\partial #1}{\partial #2}}

\newcommand{\mean}{\mathbb{E}}
\newcommand{\var}{{\rm I\kern-.3em D}}

\newcommand{\eps}{\varepsilon}

\newtheorem*{theorem*}{Theorem}
\newtheorem*{proposition*}{Proposition}
\newtheorem*{example*}{Example}
\DeclareMathSymbol{\shortminus}{\mathbin}{AMSa}{"39}
\usepackage[many]{tcolorbox}

\newcommand{\smallerfont}{\fontsize{8}{8}\selectfont}

\usepackage[framemethod=TikZ]{mdframed}
\mdfdefinestyle{mybox}{%
    linecolor=white,
    outerlinewidth=1pt,
    roundcorner=2pt,
    innertopmargin=10pt,
    innerbottommargin=0.5pt,
    innerrightmargin=10pt,
    innerleftmargin=10pt,
    backgroundcolor=black!3!white}

\tcbuselibrary{theorems}

\newtcbtheorem[number within=section]{mytheo}{Statement}%
{colback=green!5,colframe=green!35!black,fonttitle=\bfseries}{th}

\newtcbox{\hlroundedpurple}[1][PastelPurpleLight]{ 
  on line,
  arc=4pt, 
  colback=#1,
  colframe=#1,
  boxrule=0pt,
  boxsep=0pt,
  left=1pt, 
  right=1pt, 
  top=2pt, 
  bottom=2pt, 
  leftrule=0pt,
  rightrule=0pt,
  toprule=0pt,
  bottomrule=0pt,
}

\newtcbox{\hlroundedgreen}[1][PastelGreenLight]{ 
  on line,
  arc=4pt, 
  colback=#1,
  colframe=#1,
  boxrule=0pt,
  boxsep=0pt,
  left=1pt, 
  right=1pt, 
  top=2pt, 
  bottom=2pt, 
  leftrule=0pt,
  rightrule=0pt,
  toprule=0pt,
  bottomrule=0pt,
}

\newtcbox{\hlrounded}[1][PastelPurpleLight]{ 
  arc=4pt, 
  colback=#1,
  colframe=#1,
  boxrule=0pt,
  boxsep=0pt,
  left=1pt, 
  right=1pt, 
  top=2pt, 
  bottom=2pt, 
  leftrule=5pt,
  rightrule=5pt,
  toprule=5pt,
  bottomrule=5pt,
}

\makeatletter
\let\save@mathaccent\mathaccent
\newcommand*\if@single[3]{%
  \setbox0\hbox{${\mathaccent"0362{#1}}^H$}%
  \setbox2\hbox{${\mathaccent"0362{\kern0pt#1}}^H$}%
  \ifdim\ht0=\ht2 #3\else #2\fi
  }
\newcommand*\rel@kern[1]{\kern#1\dimexpr\macc@kerna}
\newcommand*\widebar[1]{\@ifnextchar^{{\wide@bar{#1}{0}}}{\wide@bar{#1}{1}}}
\newcommand*\wide@bar[2]{\if@single{#1}{\wide@bar@{#1}{#2}{1}}{\wide@bar@{#1}{#2}{2}}}
\newcommand*\wide@bar@[3]{%
  \begingroup
  \def\mathaccent##1##2{%
    \let\mathaccent\save@mathaccent
    \if#32 \let\macc@nucleus\first@char \fi
    \setbox\z@\hbox{$\macc@style{\macc@nucleus}_{}$}%
    \setbox\tw@\hbox{$\macc@style{\macc@nucleus}{}_{}$}%
    \dimen@\wd\tw@
    \advance\dimen@-\wd\z@
    \divide\dimen@ 3
    \@tempdima\wd\tw@
    \advance\@tempdima-\scriptspace
    \divide\@tempdima 10
    \advance\dimen@-\@tempdima
    \ifdim\dimen@>\z@ \dimen@0pt\fi
    \rel@kern{0.6}\kern-\dimen@
    \if#31
      \overline{\rel@kern{-0.6}\kern\dimen@\macc@nucleus\rel@kern{0.4}\kern\dimen@}%
      \advance\dimen@0.4\dimexpr\macc@kerna
      \let\final@kern#2%
      \ifdim\dimen@<\z@ \let\final@kern1\fi
      \if\final@kern1 \kern-\dimen@\fi
    \else
      \overline{\rel@kern{-0.6}\kern\dimen@#1}%
    \fi
  }%
  \macc@depth\@ne
  \let\math@bgroup\@empty \let\math@egroup\macc@set@skewchar
  \mathsurround\z@ \frozen@everymath{\mathgroup\macc@group\relax}%
  \macc@set@skewchar\relax
  \let\mathaccentV\macc@nested@a
  \if#31
    \macc@nested@a\relax111{#1}%
  \else
    \def\gobble@till@marker##1\endmarker{}%
    \futurelet\first@char\gobble@till@marker#1\endmarker
    \ifcat\noexpand\first@char A\else
      \def\first@char{}%
    \fi
    \macc@nested@a\relax111{\first@char}%
  \fi
  \endgroup
}
\makeatother

\usepackage{fancyhdr}
\usepackage[utf8]{inputenc} 
\usepackage[T1]{fontenc}    
\usepackage{hyperref}       
\usepackage{url}            
\usepackage{booktabs}       
\usepackage{amsfonts}       
\usepackage{nicefrac}       
\usepackage{microtype}      
\usepackage{tabularx}
\usepackage{thm-restate}
\usepackage[ruled,vlined]{algorithm2e}

\title{Self-Refining Training for\\ Amortized Density Functional Theory}

\author{%
  Majdi Hassan\thanks{Correspondence: \texttt{\href{mailto:majdi.mhas@gmail.com}{majdi.mhas@gmail.com} (Majdi Hassan)}}\,\,~$^{1,2}$
  \And
  Cristian Gabellini\,\,~$^{3}$
  \And
  Hatem Helal\,\,~$^{3}$
  \AND
  Dominique Beaini\,\,~$^{1,2,3}$
  \And
  Kirill Neklyudov\,\,~$^{1,2}$
  \AND
  \textnormal{
  $^{1}$Université de Montréal \quad
  $^{2}$Mila - Quebec AI Institute \quad
  $^{3}$Valence Labs}
}

\begin{document}

\maketitle

\begin{abstract}
Density Functional Theory (DFT) allows for predicting all the chemical and physical properties of molecular systems from first principles by finding an approximate solution to the many-body Schrödinger equation. However, the cost of these predictions becomes infeasible when increasing the scale of the energy evaluations, e.g., when calculating the ground-state energy for simulating molecular dynamics. Recent works have demonstrated that, for substantially large datasets of molecular conformations, Deep Learning-based models can predict the outputs of the classical DFT solvers by amortizing the corresponding optimization problems. In this paper, we propose a novel method that reduces the dependency of amortized DFT solvers on large pre-collected datasets by introducing a self-refining training strategy. Namely, we propose an efficient method that simultaneously trains a deep-learning model to predict the DFT outputs and samples molecular conformations that are used as training data for the model. We derive our method as a minimization of the variational upper bound on the KL-divergence measuring the discrepancy between the generated samples and the target Boltzmann distribution defined by the ground state energy. To demonstrate the utility of the proposed scheme, we perform an extensive empirical study comparing it with the models trained on the pre-collected datasets. Finally, we open-source our implementation of the proposed algorithm, optimized with asynchronous training and sampling stages, which enables simultaneous sampling and training. Code is available at \href{https://github.com/majhas/self-refining-dft}{https://github.com/majhas/self-refining-dft}.
\end{abstract}

\section{Introduction}
\label{sec:intro}

\begin{figure}
    \vspace{-30pt}
    \centering
    \includegraphics[width=\textwidth, trim=2cm 1.5cm 2cm 1cm, clip]{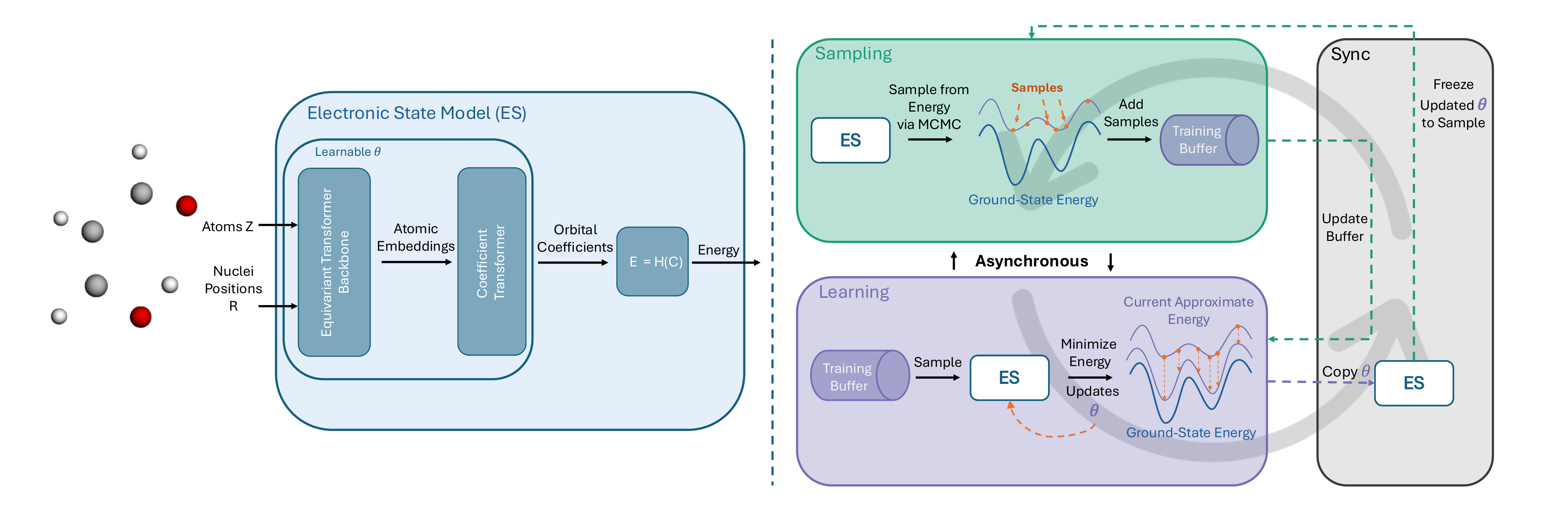}
    \caption{
    \small Overview of the self-refining training. From atom types $Z$ and positions $R$, the electronic state model predicts orbital coefficients which are used to compute the energy. Self-Refining Training operates in two asynchronous phases: (1) \sampling conformations under the current energy estimation, and (2) \learning of the sampled conformations updates the model parameters $\theta$. A synchronization step periodically updates the parameters of the sampler with the latest parameters from the \learning phase.}
    \label{fig:closed_loop}
    \vspace{-15pt}
\end{figure}
\looseness=-1
Density Functional Theory (DFT) is a quantum-mechanical approximation for solving the electronic structure problem \citep{hohenberg_inhomogeneous_1964, kohn_self-consistent_1965, roothaan_new_1951}, commonly used in drug design, materials discovery, and the characterization of chemical properties \citep{pribram-jones_dft_2015}. 
Among various computational methods, DFT is favored for its balance of accuracy and scalability \citep{keith_combining_2021}. However, conventional DFT solvers usually focus on a single Hamiltonian, i.e. single geometry of the given molecule \citep{parr_density_1980}. This naturally makes them unsuitable for tasks that require sequentially evaluating energy across multiple configurations, such as conformational sampling, transition path sampling, and molecular dynamics \citep{leimkuhler2015molecular}. In these tasks, for every generated geometry, one has to re-run DFT computations from scratch to find the electron density that minimizes the total energy \citep{payne_iterative_1992}.

\looseness=-1
In contrast, deep learning (DL) is a promising approach to scaling DFT-based computations, e.g. predicting energies to near quantum-chemical accuracy or determining the converged Hamiltonian matrix. In general, deep learning methods offer excellent scalability and generalizability, provided sufficiently large, high-quality datasets are available \citep{deng_imagenet_2009, beaini_towards_2023}. Yet, the creation of these datasets presents a major bottleneck because the labeling process involves running the conventional DFT methods (e.g. self-consistent field (SCF) method \citep{roothaan_new_1951}) for each data point. This process makes data generation prohibitively expensive at large scales. For instance, it took \emph{two years} to generate 3.8 million conformations for the PCQ dataset \citep{nakata_pubchemqc_2017}.

\looseness=-1
Recent work \citep{zhang_self-consistency_2024, mathiasen_reducing_2024, li_neural-network_2024} has taken a step toward overcoming this data-labeling burden by proposing a different training scheme. Namely, instead of predicting the energy, they predict the electronic state and define \textit{implicit DFT loss} as the energy of the predicted electronic state, which amounts to a single iteration of SCF compared to a hundred required for labeling. We call this family of models as electronic state (ES) models. 
This approach demonstrated comparable accuracy to prior work \citep{schutt_unifying_2019, unke_machine_2021, yu_efficient_2023} while avoiding the dependency on data labeling.
In our approach, we push this idea further by reducing the need for pre-generating the molecular conformations themselves, thus enabling the self-refining training of the ES model where we simultaneously train the electronic state model to minimize the energy via the implicit DFT loss and sample conformations to be used as training data.



\looseness=-1
\paragraph{Main contributions} We propose a framework for learning the electronic state model that accurately predicts DFT outputs while significantly reducing the reliance on pre-collected data. The key idea is that the data generation and the energy estimation steps are complementary problems to each other, i.e. a partial solution of one step enhances the performance of another, creating a virtuous cycle. Thus, one can use the current (in the process of training) ES model to generate new samples (\sampling step), which are then used for further training of the ES model (\learning step). The proposed framework rests on the fact that the minimum of the predicted electronic state energy has to be achieved \textit{point-wise} which allows for a great flexibility in the choice of the distribution of samples. We formalize this insight by deriving a variational upper bound on the KL-divergence between the distribution of samples and the target Boltzmann distribution of the ground-state energy. Furthermore, we demonstrate that the Wasserstein gradient \citep{ambrosio_gradient_2007} of the derived upper bound corresponds to the Langevin dynamics, which motivates its usage for generating new samples. 

\looseness=-1
We empirically compare our method with the baselines and demonstrate that it achieves better performance on the test-set when using the same amount of the training data and superior performance on samples using the overdamped molecular dynamics. These properties are especially interesting in the data-scarce scenarios, which we highlight by achieving chemical accuracy using as few as $25$ conformations. Lastly, we provide an open-source implementation of our method that allows for efficient asynchronous training.

\section{Background}
\label{sec:background}

\looseness=-1
\subsection{The Schrödinger Equation}
\label{sec:scroedinger_eq}
\looseness=-1
Quantum mechanics offers a complete theory for predicting all the chemical and physical properties of atomic systems through the solution of the stationary Schrödinger equation.
Namely, for the given Hamiltonian operator $\hat{H}$ acting on the wavefunctions $\psi(r_1,\ldots,r_n):\mathbb{R}^{3n} \to \mathbb{C}$ of $n$ electrons at positions $r_1,\ldots,r_n$, one has to find all the eigenvalues $E_i$ and the corresponding eigenstates $\psi_i$, i.e.
\begin{align}
    \hat{H} \psi_i(r_1,\ldots,r_n) = E_i \psi_i(r_1,\ldots,r_n)\,.
    \label{eq:stationary_schrod}
\end{align}
For molecule with nuclei at positions $R$ and atomic charges $Z$, the Hamiltonian operator is
\begin{align}
    \hat{H} = -\frac{1}{2}\sum_{i=1}^n\nabla^2_i + \sum_{i < j} \frac{1}{\norm{r_i - r_j}} - \sum_{i,I} \frac{Z_I}{\norm{r_i - R_I}}  + \sum_{I < J} \frac{Z_IZ_J}{\norm{R_I - R_J}}\,,
\end{align}
where $\nabla^2_i\psi = \inner{\nabla_i}{\nabla_i\psi}$ and $\nabla_i$ is the gradient w.r.t. the position $r_i$ of the $i$-th electron. This is the so-called Born-Oppenheimer approximation where the positions of the nuclei $R$ are fixed input parameters of Hamiltonian and the operator acts on wavefunctions $\psi$ which define electronic states and have to respect the fermionic anti-symmetry, i.e. $\psi(r_i,\ldots,r_j) = -\psi(r_j,\ldots,r_i)$

\looseness=-1
For many applications, the eigenstate $\psi_0$ (\textit{the ground state}) corresponding to the lowest eigenvalue $E_0$ (the lowest energy) is of a particular interest.
Alternatively to \cref{eq:stationary_schrod}, this state can be defined as the solution of the following variational problem
\begin{align}
    E_0 = \min_\psi \frac{\inner{\psi}{\hat{H}\psi}}{\inner{\psi}{\psi}}\,, \;\;\inner{\psi}{\xi}\coloneqq\int dr_1,\ldots,dr_n\; \psi^*(r_1,\ldots,r_n)\xi(r_1,\ldots,r_n)\,,
    \label{eq:wf_minimization}
\end{align}
where $\psi^*$ is the complex conjugate of $\psi$. Finding $\psi_0(r_1,\ldots,r_n)$ by directly minimizing the energy is an extremely challenging computational problem because it requires solving the optimization problem w.r.t. high-dimensional functions $\psi:\mathbb{R}^{3n} \to \mathbb{C}$.

\looseness=-1
\subsection{Density Functional Theory}
\label{sec:dft}
\looseness=-1
As an attempt to avoid the optimization in the space of high-dimensional wavefunctions \cite{hohenberg1964inhomogeneous} established Density Functional Theory (DFT), which proposes an alternative formulation to the variational problem in \cref{eq:wf_minimization}. Namely, instead of the optimization in the space of wavefunctions ($\psi: \mathbb{R}^{3n} \to \C$), one can minimize energy w.r.t. the electronic density ($\rho: \mathbb{R}^{3} \to \R$)
\begin{align}
    \label{eq:elec_density}
    E_0 = \min_\psi \frac{\inner{\psi}{\hat{H}\psi}}{\inner{\psi}{\psi}} = \min_\rho F[\rho] + V[\rho]\,,\;\;\rho(r) = \int dr_2\ldots dr_n\; |\psi(r,r_2,\ldots,r_n)|^2\,,
\end{align}
where $\rho(r)$ is the normalized electronic density (i.e. $\int dr\rho(r) = n$), $F[\rho]$ is the universal density functional represented by the parts of the Hamiltonian that do not depend on $R$, and $V[\rho]$ is the system-dependent functional that is defined by nuclei positions $R$. However, the analytic form of the universal functional $F[\rho]$ is unknown and its efficient approximation is the main challenge in DFT.

\looseness=-1
The next conceptual leap forward was made by ``simulating'' many-body electron interactions via the effective one-electron potential \citep{Kohn65}, which allowed for reducing the problem of $n$ interacting electrons to $n$ independent one-electron problems. The latter can be efficiently solved under the Hartree-Fock family of the wavefunctions \citep{roothaan_new_1951}. Namely, one can parameterize the wavefunction using the orthonormal set of molecular orbitals $\psi_i$ as follows
\begin{align}
    \psi(r_1,\ldots,r_n) = \det \left[\Psi_{HF}\right]\,,\;\; (\Psi_{HF})_{ij} = \psi_i(r_j)\,,\;\; \inner{\psi_i}{\psi_j} = \delta_{ij}\,.
\end{align}
Parameterizing the wavefunction as the determinant of $\Psi_{HF}$ guarantees the fermionic anti-symmetry.

\looseness=-1
Linear Combination of Atomic Orbitals (LCAO) is a common approach for parameterizing molecular orbitals $\psi_i(r)$. Namely, each single-particle wavefunction is expanded in a predefined basis set, typically localized around atomic centers $\psi_i(r) \;=\; \sum_{\mu} c_{\mu i}\,\phi_\mu(r-R_\mu)$, where $c_{\mu i}$ are the coefficients of $\psi_i(r)$ in the new basis $\phi_\mu(r-R_\mu)$ \citep{hehre_selfconsistent_1969}. Variational problem in \cref{eq:elec_density} becomes
\begin{equation}
    \label{eq:dft_energy}
    E_0 = \min_C E(C)\,, \;\text{ subject to }\; C^\dagger SC = \one\,,\;\; S_{\mu \nu} = \int dr\; \phi_\mu^*(r - R_\mu) \phi_\nu(r - R_\nu)\,,
\end{equation}
where $C$ is the matrix of coefficients $c_{\mu i}$ in the new basis, $E(C)$ is the energy functional in this basis, and $CSC = \one$ is the orthonormality condition of the orbitals $\psi_i$. Significant progress has already been made in the design of the effective potential and the basis set. That is why, throughout the paper, we assume that given the nuclei positions $R$ we know the analytic form of the energy functional $E(C)$ and the overlap matrix $S$.



\section{Self-Refining Training for Amortized DFT}
\label{sec:method}

In this section, we present our method that simultaneously learns the DFT-based energy model of molecular conformations and samples these conformations from the Boltzmann distribution of the learned energy model. The foundation of our method is the variational upper bound on the KL-divergence between generated samples and the true Boltzmann density of molecular conformations which we derive in \cref{sec:upper_bound}. Minimization of this upper bound w.r.t.\ the parameterized electronic state model corresponds to minimization of the predicted ground state energy on the samples (see \cref{sec:energymin}). Whereas the minimization of the upper bound w.r.t.\ the distribution of the samples results in sampling from the Boltzmann density corresponding to the \textit{parameterized} energy model (see \cref{sec:sampling}). Building on these results, we propose Self-Refining Training algorithm in \cref{sec:algorithm}.

\subsection{Variational Energy Bound}
\label{sec:upper_bound}
\begin{algorithm}[t]
    \caption{Self-Refining Training pseudocode (for \sampling and \learning)}
    \label{alg:general}
    \SetKwInOut{Input}{Input}
    \SetNoFillComment
    \DontPrintSemicolon

    \Input{Parametric model $f_\theta$ for which $f_\theta(R)^\dagger S(R)f_\theta(R) = \one$, the energy functional $E(R,C)$.}
    
    \While{not converged}{
        \hlroundedgreen{$\sample R \sim q_\theta(R) \propto \exp{( - E(R, f_\theta(R)) )}$} \tcp*[r]{\smallerfont using Monte Carlo methods}
        \hlroundedpurple{$\update f_\theta$ by minimizing $\mean_{q(R)} \bigl[E(R, f_\theta(R))\bigr]$} \tcp*[r]{\smallerfont using the collected samples $R$}
    }
    \Return{$\theta$}
\end{algorithm}

\looseness=-1
Deep Learning allows for an efficient amortization of the optimization problem from \cref{eq:dft_energy} over different molecular geometries defined by the nuclei positions $R$ \citep{zhang_self-consistency_2024,mathiasen_reducing_2024}. Indeed, making the energy functional $E(C)$ and the overlap matrix $S$ explicitly dependent on $R$, one can define the energy model $E_0(R)$ as follows
\begin{equation}
    \label{eq:amortized_dft}
    E_0(R) = \min_C E(R, C), \;\text{ subject to }\; C^\dagger S(R)C = \one.
\end{equation}
However, instead of directly modeling $E_0(R)$, we define \textit{the electronic state model} as the solution of the following optimization problem
\begin{align}
    f^*(R) = \argmin_{C:\ C^\dagger S(R)C = \one} E(R, C)\,, \text{ which defines the energy model } E_0(R) = E(R, f^*(R))\,.
\end{align}
The target distribution of molecular conformations is defined by the following Boltzmann distribution
\begin{align}
    p^*(R) = \frac{1}{Z^*}\exp\left(-E(R, f^*(R))\right)\,, \;\; Z^* = \int dR\; \exp\left(-E(R, f^*(R))\right)\,.
\end{align}
However, in practice, we have access neither to the density $p^*(R)$ nor to the samples from this density. The only supervision signal we have is the functional $E(R, C)$ from the optimization problem in \cref{eq:amortized_dft}, which defines the energy. In order to collect samples from $p^*(R)$ and learn its density model, we derive a practical upper bound on the KL-divergence between any density $q(R)$ and $p^*(R)$ in the following proposition.
\begin{mdframed}[style=MyFrame2]
\begin{restatable}{proposition}{upperbound}
    \label{prop:upper_bound}
    \textup{[Variational Energy Bound]}
    For any density $q(R)$, the KL-divergence between $q(R)$ and the ground-state Boltzmann distribution $p^*(R) \propto\exp(-E(R, f^*(R)))$ admits the following variational upper bound
    \begin{align} 
    \KL(q,p^*) \leq \underbrace{\mean_{q(R)} \log q(R) + \mean_{q(R)} E(R,f_\theta(R))}_{\coloneqq \Phi[q,f_\theta]} + \log Z^*\,,
    \label{eq:upper_bound}
    \end{align}
    where $f_\theta$ is a parametric approximation of $f^*$, which satisfies $f_\theta(R)^\dagger S(R)f_\theta(R) = \one$. For any $q(R)$, the bound becomes tight when $f_\theta(R) = f^*(R)$.
\end{restatable}
\end{mdframed}
See \cref{app:proof_ub} for the proof. 

\looseness=-1
Both steps of our algorithm can be described as minimization of the upper bound in \cref{prop:upper_bound}: the \sampling step corresponds to the minimization of $\Phi[q,f_\theta]$ w.r.t. the distribution of samples $q(R)$ and the \learning step corresponds to the minimization of $\Phi[q,f_\theta]$ w.r.t. the parameters of the electronic state model $\theta$. In \cref{alg:general}, we provide pseudocode for both steps in the most general form. Notably, these steps can be run asynchronously with periodic synchronization for exchanging the newly generated samples and current parameters of the electronic state model.

\subsection{Energy Minimization}
\label{sec:energymin}
\looseness=-1
The \learning step corresponds to the learning the electronic state model $f_\theta(R)$ by minimization of the upper bound from \cref{prop:upper_bound}. This can be done by minimizing $\Phi[q,f_\theta]$ w.r.t. the parameters of the model $\theta$. However, first, we have to parameterize $f_\theta$ that satisfies the orthonormality conditions $f_\theta(R)^\dagger S(R)f_\theta(R) = \one$. To handle this constraint, we follow \cite{head-gordon_optimization_1988, kasim_dqc_2022, li_d4ft_2023} and define the electronic state model as
\begin{align}\label{eq:reparam}
    f_\theta(R) = U\Lambda^{-1/2}U^T Q_\theta(R)\,,\;\; Q_\theta(R) = \texttt{Orthogonal}(\texttt{NN}_\theta(R))\,,
\end{align}
where $S(R) = U\Lambda U^T$, $U$ and $\Lambda$ are the corresponding matrices of eigenvectors and eigenvalues of $S(R)$, $\texttt{NN}_\theta(R)$ is the neural network output for the input coordinates $R$, and \texttt{Orthogonal} is any orthogonolization transformation, e.g. the QR-decomposition or the Cayley transform (in practice, we use the QR-decomposition). Thus, by straightforward calculations we have $f_\theta(R)^\dagger S(R)f_\theta(R) = \one$.

\looseness=-1
The minimization of the upper bound $\Phi[q,f_\theta] + \log Z^*$ boils down to the minimization of the energy model $E(R,f_\theta(R))$ on the samples from $q(R)$. We formalize this in the following statement, which is a straightforward corollary of \cref{prop:upper_bound}.
\begin{mdframed}[style=MyFrame2, backgroundcolor=PastelPurpleLight]
\begin{restatable}{corollary}{energymin}
    \label{prop:energymin}
    \textup{[Energy Minimization]}
    For any density $q(R)$ which is positive $q(R) > 0$ for the states with finite energies $\forall\; R: E(R,f^*(R)) < \infty$ except for sets of zero measure, we have
    \begin{align}
        f^* = \argmin_{f_\theta} \Phi[q,f_\theta] = \argmin_{f_\theta} \mean_{q(R)}E(R,f_\theta(R))\,.
        \label{eq:energy_minimization}
    \end{align}
\end{restatable}
\end{mdframed}
The practical value of this corollary is that the optimization of the electronic state model $f_\theta$ does not require any labeled data like ground truth energies or precalculated solutions of the optimization problem \cref{eq:amortized_dft}. This makes our framework independent of expensive dataset collection, analogously to the recent works by \cite{zhang_self-consistency_2024, mathiasen_reducing_2024, li_neural-network_2024}. Indeed, the objective in \cref{eq:energy_minimization} can be efficiently optimized using gradient-based approaches (we provide pseudocode in \cref{alg:cl}).

\subsection{Sampling}
\label{sec:sampling}
\looseness=-1
Optimization of the upper bound from \cref{prop:upper_bound} w.r.t. the distribution $q(R)$ results in the \sampling step. Namely, for the fixed parameteric model $f_\theta(R)$, in the following corollary we find the distribution $q(R)$ that minimizes $\Phi[q, f_\theta]$. 

\begin{mdframed}[style=MyFrame2, backgroundcolor=PastelGreenLight]
\begin{restatable}{corollary}{samplingboltz}
    \label{prop:samplingboltz}
    \textup{[Sampling]}
    For any electronic state model $f_\theta(R)$, the optimum of the upper bound $\Phi[q,f_\theta]$ w.r.t. the density $q(R)$ is the Boltzmann distribution defined by the model $f_\theta(R)$, i.e.
    \begin{align}
        q^*(R) = \argmin_{q}\Phi[q,f_\theta] = \frac{1}{Z_\theta}\exp(-E(R,f_\theta(R)))\,.
        \label{eq:target_boltz}
    \end{align}
\end{restatable}
\end{mdframed}
See \cref{app:proof_samplingboltz} for proof. This result suggests that given the current approximation of the electronic state model $f_\theta$ the best distribution $q(R)$ we can get is the Boltzmann distribution corresponding to the current energy model $E(R,f_\theta(R))$.

\looseness=-1
Reducing the problem to sampling from the unnormalized density opens the door to a rich field of Monte Carlo methods, among which we choose the Langevin dynamics, which is motivated by the variational bound proposed in \cref{prop:upper_bound}. Indeed, sampling from the Boltzmann density from \cref{eq:target_boltz} is different from the classical sampling problems because it happens in parallel to the updates of the parameters $\theta$ and the density itself. In this context, tuning an algorithm or training a model to sample from the target density might be wasteful if the density changes too quickly. 

\looseness=-1
Instead, we assume that at every iteration the distribution $q(R)$ is reasonably close to the target $\exp(-E(R,f_\theta(R)))$. Therefore, we need to make several updates of $q(R)$ that minimize the bound $\Phi[q,f_\theta]$. This can be formalized as following the negative gradient of $\Phi[q,f_\theta]$ w.r.t. the density $q(R)$. The natural choice for the gradient in the space of distribution is the Wasserstein gradient \citep{ambrosio_gradient_2007} since it can be efficiently implemented on the sample-level as transporting the samples along a vector field. We derive the Wasserstein gradient dynamics minimizing $\Phi[q,f_\theta]$ in the following proposition.

\begin{mdframed}[style=MyFrame2, backgroundcolor=PastelGreenLight]
\begin{restatable}{proposition}{wgradient}
    \label{prop:wgradient}
    \textup{[Wasserstein Gradient]}
    For the functional $\Phi[q,f_\theta]$ defined in \cref{prop:upper_bound}, the gradient descent w.r.t. $q(R)$ corresponds to the following PDE
    \begin{align}
        \deriv{q_t(R)}{t} = -\inner{\nabla_R}{q_t(R)\left(-\nabla_R E(R,f_\theta(R))\right)} + \Delta_R q_t(R)\,,
    \end{align}
    which can be efficiently simulated via the following Stochastic Differential Equation (SDE)
    \begin{align}
        dR_t = -\nabla_R E(R_t,f_\theta(R_t))dt + \sqrt{2}dW_t\,,
    \end{align}
    where $W_t$ is the standard Wiener process.
\end{restatable}
\end{mdframed}
For the proof see \cref{app:proof_wgradient}. This results suggests generating new samples via the Langevin dynamics whereas the gradient $\nabla_R E(R_t,f_\theta(R_t))$ can be efficiently calculated by backpropagation (see \cref{alg:cl} for the pseudocode).

\subsection{Self-Refining Training Algorithm}
\label{sec:algorithm}

\begin{algorithm}[t]
    \caption{Self-Refining Training Algorithm}
    \label{alg:cl}
    \SetKwInOut{Input}{Input}
    \SetNoFillComment
    \DontPrintSemicolon


    \Input{Electronic state model $f_\theta$, energy functional $E(R,C)$, replay buffer $\cB$, length of the sampling chain $T$, step size $dt$, initial distribution of samples $p_0(R)$ and probability $P$} 
    
    \While{not converged}{
        \hlroundedgreen{$\sample$ from current approximate of energy $E(R, f_\theta(R))$}
        \BlankLine
        $R_0 \sim \cB$ with prob. $P$ or $R_0 \sim p_0(R)$ with prob. $(1-P)$ \tcp*[r]{\smallerfont sample initial point}
        \BlankLine
        \For{$t=0, \dots, T-dt$}{
            $R_{t+dt} \gets R_{t} - \nabla_R E(R_t, f_\theta(R_t)) dt + \sqrt{2} dW_t$ \tcp*[r]{\smallerfont Langevin dynamics}
        }
        $\cB \leftarrow \cB \cup R_T$ \tcp*[r]{\smallerfont update the buffer $\cB$}
        
        \hlroundedpurple{$\update$ electronic state model using samples from $\cB$}
        \BlankLine

        $\texttt{grad}_\theta \gets \nabla_\theta \mean_{R\sim \cB} E(R, f_\theta(R))$ \tcp*[r]{\smallerfont estimate the gradient using samples from $\cB$}
        \BlankLine
        
        $\theta \leftarrow \texttt{Optimizer}(\theta, \texttt{grad}_\theta)$ \tcp*[r]{\smallerfont minimize the energy using the gradient}
    }
    \Return{$\theta$}
\end{algorithm}

The \learning and \sampling steps together result in the Self-Refining Training algorithm, which we describe in \cref{alg:cl}. The essential part of the algorithm that allows for self-refinement is the energy functional $E(R,C)$ that provides the supervision signal in every point of the space $R,C$. This is similar but not identical to adversarial training \citep{goodfellow2014explaining} where only small perturbations of the training data are allowed, otherwise one has to re-label the generated sample. Furthermore, training on the samples from the Langevin dynamics aligns with the downstream applications of the learned energy model since it corresponds to the overdamped molecular dynamics.

In theory, Self-Refining Training does not require any pre-collected data, but, in practice, we find that pre-training the ES model $f_\theta(R)$ even on a small number of conformations provides a better initialization for further training. Additionally, for the initial distribution $p_0(R)$ of the Langevin dynamics we use the conformations from either the replay buffer or from the dataset. This provides the model with both ``clean'' and ``noisy'' samples for the learning stage, which improves the model's ability to generalize as we will demonstrate later in the experiments section. Further details can be found in \cref{appendix:train_details}.

\paragraph{Asynchronous Training} Crucially, our algorithm allows for asynchronous implementation of \learning and \sampling steps. Indeed, both steps minimize the variational lower bound and their iterations resemble the coordinate descent w.r.t.\ different arguments of $E(R,C)$ (unlike adversarial networks \citep{goodfellow2014generative}). Thus, one can generate new samples and update ES model parameters independently on different sets of GPUs. In particular, samples generated by the sampling process go into a temporary buffer, which, once full, triggers a synchronization step. During this step, the new samples are added to the main replay buffer, and the sampling process is updated with the latest parameters from the learning process.

\section{Experiments}
\label{experiemnts}

\begin{table}[!t]
    \vspace{-10pt}
    \caption{
    Performance across metrics for self-refining training in the \emph{data-scarce} ($10\%$ of the available data) against a model with access to the full dataset (MD17 Hamiltonian). The ES + I-DFT is a model trained on pre-collected conformations, but using implicit DFT loss. ES + SR is a model trained with self-refining training. $\mathcal{D}_{X\%}$ means the model was trained using Self-Refinement and X\% of pre-collected data.
    The evaluations were done on the test split of conformations of each molecule from MD17. We also include regression-based models to provide a comparison to the I-DFT loss objective.
    }
    \vspace{0.1 cm}
    \centering
    \small
    \setlength\tabcolsep{1.5pt}
        \begin{tabular}{llcccccc}
        \toprule
        Molecule & Setting & $E[\mu E_\mathrm{h}]$ & $\mathbf{H}\,[\mu E_\mathrm{h}]\downarrow$ & $\epsilon [\mu E_\mathrm{h}]\downarrow$ & $\epsilon_{\mathrm{HOMO}}$$\,[\mu E_\mathrm{h}]\downarrow$ & $\epsilon_{\mathrm{LUMO}}$$\,[\mu E_\mathrm{h}]\downarrow$ & $\epsilon_\Delta$$\,[\mu E_\mathrm{h}]\downarrow$ \\
        \midrule
        \multirow{2}{*}{Ethanol} 
                & I-DFT + $\mathcal{D}_{100\%}$ & \colorcell  $63.33$  &  $91.01$ &  $413.25$ &   $489.96$ &   $1792.90$ &  $1341.80$ \\
                & I-DFT + SR  + $\mathcal{D}_{10\%}$ & $135.94$   & $117.23$  & $397.89$   & $432.88$ & $1567.50$ & $1205.20$ \\
                & I-DFT + SR + $\mathcal{D}_{100\%}$ &  $83.84$ & \colorcell  $86.36$ & \colorcell $291.74$ &  \colorcell $336.17$ &  \colorcell $1389.40$ &   \colorcell $1090.90$  \\
        \midrule
        Malondi-   & I-DFT + $\mathcal{D}_{100\%}$ &  \colorcell $773.00$ & \colorcell $165.30$ & \colorcell $655.20$ & \colorcell $622.00$   &                             \colorcell $3042.00$ & \colorcell $2679.00$ \\
        aldehyde   & I-DFT + SR  + $\mathcal{D}_{10\%}$ & $7089.00$   &  $555.50$  & $2007.00$   & $1834.00$ & $7275.00$ & $6219.00$ \\
                   & I-DFT + SR  + $\mathcal{D}_{100\%}$ & $5411.00$   & $527.00$  & $1735.00$   & $1682.00$ & $6338.00$ & $5284.00$ \\
        \midrule
        \multirow{2}{*}{Uracil} 
                & I-DFT + $\mathcal{D}_{100\%}$ &  $49.24$ & $47.05$   & $366.78$   &  $231.44$ & $1082.40$ & $1146.10$ \\
                 & I-DFT + SR  + $\mathcal{D}_{10\%}$ & $40.91$   &  $35.61$  & $220.97$   & $149.91$ & \colorcell$625.02$ & \colorcell $660.45$ \\
                 & I-DFT + SR  + $\mathcal{D}_{100\%}$ & \colorcell $38.34$   & \colorcell $33.64$  & \colorcell $205.14$   & \colorcell $142.99$ & $694.23$ & $748.30$ \\
        \midrule
        \bottomrule
        \end{tabular}
    \label{tab:res-md17}
    \vspace{-15pt}
\end{table}


\textbf{Datasets.}
We use the MD17 dataset, which contains conformations for ethanol, malondialdehyde, and uracil \cite{chmiela_machine_2017}, generated via molecular dynamics at the DFT level of theory using the def2-SVP Gaussian-Type Orbital (GTO) basis set and the PBE exchange-correlation functional \cite{perdew_generalized_1996}. We use the standard train/validation/test splits from \cite{schutt_unifying_2019}. To explore data scarcity, we consider a reduced data scenarios using only $10\%$ of the full dataset, which we run our self-refining method on top of these subsets. In the data-rich setting, we have full access to the conformations from the MD17 dataset and generate additional samples using our self-refining method.

\textbf{Metrics.} 
We follow the standard metrics introduced in prior work \cite{schutt_unifying_2019}. 
Specifically, we report the mean absolute error (MAE) between the Hamiltonian matrix derived from the model’s predicted coefficients and the reference DFT-computed Hamiltonian. 
We also measure the MAE of the orbital energies $\bm{\epsilon}$ and four molecular properties: the highest occupied molecular orbital $\epsilon_{\texttt{HOMO}}$, the lowest unoccupied molecular orbital $\epsilon_{\texttt{LUMO}}$, the energy gap $\epsilon_{\Delta}$, and the total energy $E$.

\textbf{Baselines.} 
We compare our self-refining approach to models that learn the energy via the implicit DFT loss on pre-collected conformations. All experiments are conducted on conformations from the MD17 dataset, using the 6-31G GTO basis set for ethanol and malondialdehyde, and the STO-3g GTO basis set for uracil. For all systems, we use the PBE exchange-correlation functional. Since this basis choice differs from the def2-SVP GTO basis set originally used to generate MD17, we compute ground-truth properties through the PySCF library \cite{sun_pyscf_2018} with the corresponding basis and exchange-correlation functionals, then evaluate using the metrics described above.

\subsection{Improving Performance}

We demonstrate that self-refining training improves performance both in the data-scarce scenario and on the out-of-distribution samples.



\begin{figure}
\vspace{-10pt}
    \centering
    \includegraphics[width=\linewidth]{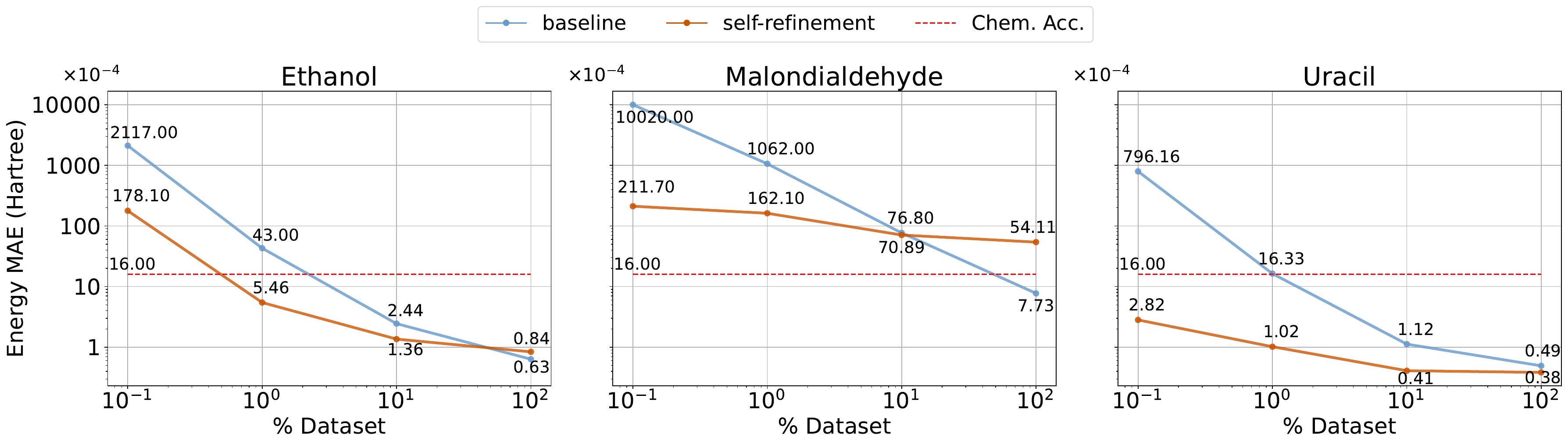}
    \caption{Energy prediction error (MAE in $\times 10^{-4}$ Hartree) of the baseline and self-refining models across varying dataset sizes for three molecules. The self-refining approach consistently improves accuracy, especially in low-data regimes, often achieving chemical accuracy (red dashed line) with significantly fewer training points.}
    \label{fig:data-perf}
\vspace{-10pt}
\end{figure}

\textbf{Data-Scarce Scenario.}
Data scarcity is a common challenge in many scientific domains where few or no conformations and their corresponding DFT labels are available. 
In these low-data regimes, data-hungry methods often perform poorly. 
To demonstrate the effectiveness of our self-refining approach, we evaluate performance under various levels of data availability -- 0.1 \%, 1\%, 10\%, 100\% -- corresponding to $25$, $250$, $2500$, $25000$ conformations, respectively, for ethanol, malondialdehyde, and uracil. 
We compare the baseline model trained using the implicit DFT loss solely on the available data against the same model augmented by the self-refining procedure. 
As shown in \cref{tab:res-md17} and \cref{fig:data-perf}, self-refinement yields performance gains in low-data settings, often matching the performance of the baseline model trained on the full dataset using only $10\%$ of the data.
These results highlight the ability of our framework to mitigate data scarcity, suggesting a viable solution for learning in data-limited settings.

\begin{figure}
    \centering
    \includegraphics[width=0.9\textwidth]{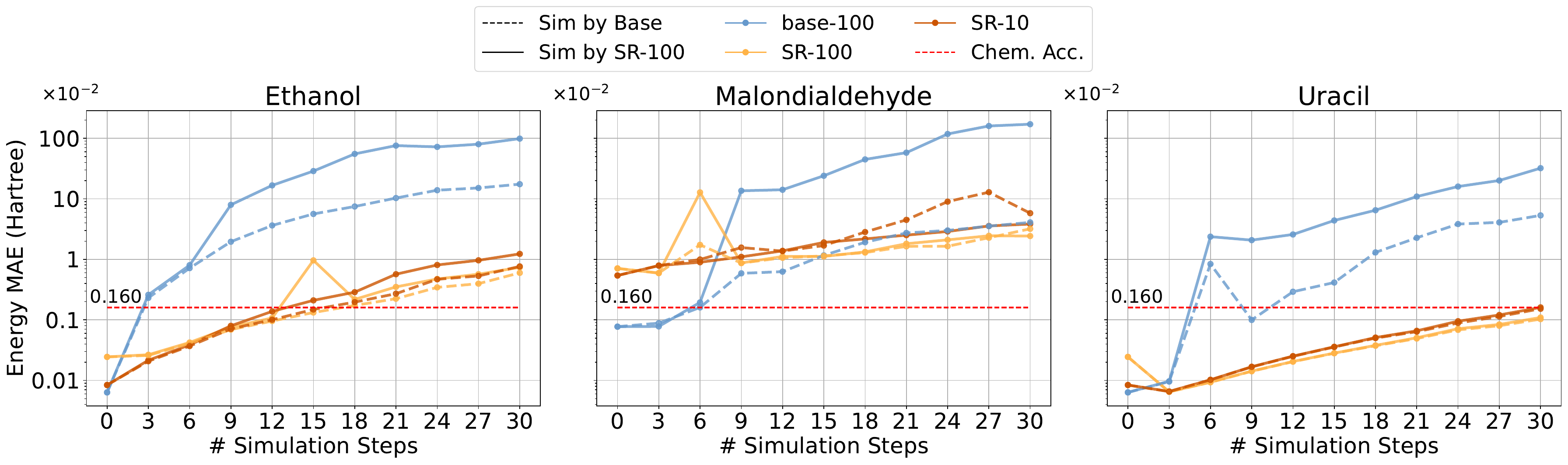}
        \label{fig:sim-perf}
    \caption{Generalization of baseline and self-refining models evaluated on samples generated from simulation. We report the energy MAE ($\times 10^{-2}$ Hartree) over 30 simulation steps for models trained on 10\% or 100\% of the data, with samples propagated by either the baseline model (dashed) or self-refining model (solid) trained on 100\% of data. The self-refining approach consistently yields lower errors and maintains accuracy further into the simulation, demonstrating improved robustness to out-of-distribution samples. The red dashed line marks the chemical accuracy threshold ($0.16 \times 10^{-2}$ Hartree).}
\vspace{-15pt}
\end{figure}

\textbf{Out-of-Distribution Scenario.}
To assess the generalization ability of our electronic state model beyond the training distribution, we evaluate the prediction accuracy of energy across simulation trajectories. 
Specifically, we consider two types of simulations: those generated by the baseline model and those generated by the self-refining model trained on 100\% of the data. 
At every 3 simulation steps, we compute the energy MAE for the baseline, SR-$10\%$, and SR-$100\%$ models. 
As shown in \cref{fig:sim-perf}, the self-refined models exhibit significantly improved robustness, maintaining low error even as the trajectory progresses further from the training distribution. 
In contrast, the baseline model quickly deteriorates, producing inaccurate energy estimates once the samples drift outside the support of the training set. 
This demonstrates that incorporating self-refinement during training leads to more reliable predictions on out-of-distribution samples.

\subsection{Efficiency from Simultaneous Training and Generation}
\begin{figure}
\vspace{-10pt}
    \centering
    \includegraphics[width=\linewidth]{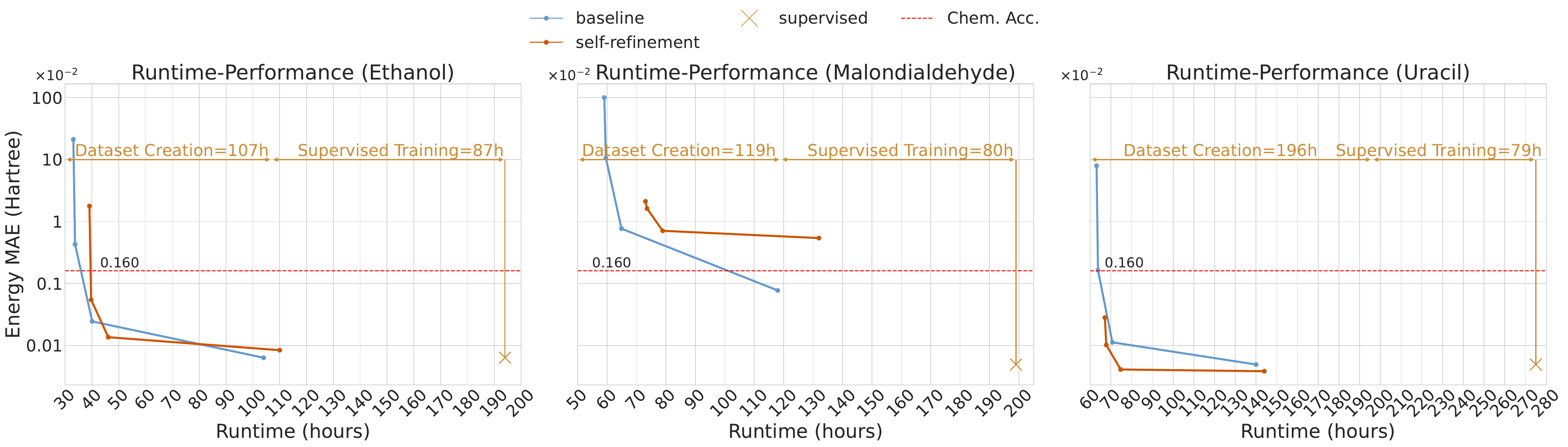}
    \caption{Runtime vs. performance comparison for baseline, self-refining, and supervised models on Ethanol and Uracil.
    We report the energy MAE ($\times 10^{-2}$ Hartree) as a function of total runtime, including data generation, labeling, and training. The self-refining method approaches the performance of supervised models with significantly lower runtime. Orange bars indicate time spent on dataset creation and supervised training for the QHNet model. The red dashed line marks the chemical accuracy threshold.}
    \label{fig:runtime-perf}
\vspace{-10pt}
\end{figure}

The self-refining method significantly reduces the overall time spent on (1) collecting conformations, (2) labeling with DFT, and (3) training the model as shown in \cref{fig:runtime-perf}, as these steps are performed simultaneously. 
A key benefit is that iterative DFT minimization need not be run to convergence for each conformation. 
Instead, amortizing the optimization by training through the implicit DFT loss, we predict ground-state solutions on the fly during inference and sample new data from an imperfect, evolving energy model. 
This capability removes the barrier of requiring fully annotated data as a prerequisite for training robust, generalizable models. Details on how the runtimes were evaluated can be found in \cref{app:runtime-analysis}. 

\section{Related Work}
\label{sec:related_works}
\looseness=-1
\paragraph{Direct Energy Minimization in DFT} Our work significantly relies on the direct energy minimization literature \citep{head-gordon_optimization_1988,kasim_dqc_2022}, which suggests finding the ground state as the state minimizing the energy of the system unlike SCF. This approach together with the frameworks for automatic differentiation  \citep{jax2018github, paszke_pytorch_2019} unlocked new family of algorithms that allows for designing stable and efficient DFT solvers \citep{weber_direct_2008, yoshikawa_automatic_2022, helal_mess_2024}. In our work, direct energy minimization naturally appears as minimization of the variational upper bound \cref{sec:energymin}.

\looseness=-1
\paragraph{Amortized DFT} Deep learning allows for training models that directly predict the optimal electronic state or energy bypassing expensive DFT solvers for every new input geometry. This can be achieved either by learning the regression models for labeled data \citep{ramakrishnan_quantum_2014, schutt_schnet_2018, falletta2024unifieddifferentiablelearningelectric}, by introducing the self-consistency loss analogous to SCF \citep{zhang_self-consistency_2024}, or, analogous to our method, by directly minimizing the energy of the predicted electronic state \citep{mathiasen_reducing_2024, li_neural-network_2024}. However, all the previous methods consider training on a fixed dataset of conformations and do not reuse the parameterized energy model for generating new ones.

\looseness=-1
\paragraph{Data-Augmentation} From the Machine Learning perspective our method can be considered as data-augmentation technique \citep{perez2017effectiveness} or adversarial training \citep{goodfellow2014explaining}. The important difference here is that our method relies neither on domain symmetry (which can be additionally utilized) nor on the existence of adversarial examples. On contrary, the Langevin dynamics that we use for generating new samples at the sampling step (see \cref{sec:sampling}) simulates the overdamped molecular dynamics; thus, we directly optimize the perfomance of our model on one of the downstream tasks.

\section{Conclusion}
\label{sec:conclusion}

In this paper, we provide a unified perspective on learning the energy model (amortized DFT) and generating conformations. The validity of our approach is established by a variational upper bound on the KL-divergence between the distribution of conformations and the ground truth Boltzmann density. Furthermore, we demonstrate that this perspective is practical and improves performance of the model across multiple systems for different available datasets.

We believe that our approach serves as an important stepping stone towards scalable and efficient \textit{ab initio} modeling of molecular systems that relies solely on the knowledge of underlying physical laws. In particular, combining the proposed algorithm with recent advances in learnable samplers \cite{noe_boltzmann_2019, akhound-sadegh_iterated_2024}, one can simultaneously train both the energy model and the generative model that samples from the corresponding Boltzmann density. Furthermore, our approach is agnostic to the implementation of the energy functional, which makes it compatible with other perspective avenues for improvement of DFT. For instance, it can be combined with stochastic estimates of the quadrature \citep{li_d4ft_2023} or orbital-free DFT calculations \citep{zhang2024overcoming, de2024leveraging}.


\section{Acknowledgments}
The research was enabled in part by computational resources provided by Mila Quebec AI Institute (https://mila.quebec) and Valence Labs / Recursion. 
KN was supported by IVADO and Institut Courtois. 

\bibliographystyle{apalike}
\bibliography{refs}
\clearpage

\appendix

\section{Proofs}
\label{appendix:proofs}

\subsection{Proof of \cref{prop:upper_bound}}
\label{app:proof_ub}

\begin{mdframed}[style=MyFrame2]
\upperbound*
\end{mdframed}
\begin{proof}
We denote the solution $f^*$ to the optimization problem in \cref{eq:amortized_dft} as
\begin{align}
    f^*(R) = \argmin_{C:\ C^\dagger S(R)C = \one} E(R, C)\,.
\end{align}
We define $f_\theta(R)$ as a parametric model for which we guarantee
\begin{align}
    f_\theta(R)^\dagger S(R)f_\theta(R) = \one\,, 
\end{align}
which can be done using the QR-decomposition (see \cref{sec:energymin} for the explicit construction). Then, for this parametric family, we have
\begin{align}
    E(R, f^*(R)) ~&\leq E(R, f_\theta(R))\,,\;\;\forall\,\theta\,,\\
    -E(R, f^*(R)) \pm \log Z^* ~&\geq -E(R, f_\theta(R)) \pm \log Z\,,\\
    \log p^*(R) + \log Z^* ~&\geq \log p(R) + \log Z\,,\\
    \mean_{q(R)}\log p^*(R) + \log Z^* ~&\geq \mean_{q(R)}\log p(R) + \log Z\,,\\
    \log \frac{Z^*}{Z} ~&\geq \mean_{q(R)}\log \frac{p(R)}{p^*(R)}\,,
\end{align}
where $p(R) \propto \exp(-E(R,f_\theta(R)))$ is the Boltzmann density of the parametrized energy model and $Z = \int dR\;\exp(-E(R,f_\theta(R)))$ is its normalization constant.

Consider the minimization of the KL-divergence between the distribution of samples $q$ and the the target Boltzmann distribution $p^*$, then we have the following estimate
\begin{align}
    \KL(q,p^*) \coloneqq \mean_{q(R)}\log \frac{q(R)}{p^*(R)} =~& \mean_{q(R)} \log \frac{q(R)}{p(R)} + \mean_{q(R)} \log \frac{p(R)}{p^*(R)}\,, \;\;\forall\, p(R)\\
    \leq~& \mean_{q(R)} \log \frac{q(R)}{p(R)} + \log \frac{Z^*}{Z} = \KL(q,p) + \log \frac{Z^*}{Z}\,,
\end{align}
which holds for any $p(R)$, but choosing $p(R) = Z^{-1}\exp(-E(R,f_\theta(R)))$, we get
\begin{align}
    \KL(q,p) + \log \frac{Z^*}{Z} =~& \mean_{q(R)} \log q(R) + \mean_{q(R)}\left(-\log p(R) - \log Z\right) + \log Z^*\\
    =~& \underbrace{\mean_{q(R)} \log q(R) + \mean_{q(R)} E(R,f_\theta(R))}_{\Phi[q,f_\theta]} + \log Z^*\,.
\end{align}
Thus, we have
\begin{align}
    \KL(q,p^*) \leq \underbrace{\mean_{q(R)} \log q(R) + \mean_{q(R)} E(R,f_\theta(R))}_{\Phi[q,f_\theta]} + \log Z^*\,.
\end{align}
Note that taking $f_\theta(R) = f^*(R)$, the upper bound becomes
\begin{align}
    \KL(q,p^*) \leq~& \mean_{q(R)} \log q(R) + \mean_{q(R)} E(R,f_\theta(R)) + \log Z^* \\
    =~& \mean_{q(R)} \log q(R) -\left[ \mean_{q(R)} (-E(R,f^*(R))) - \log Z^*\right] = \KL(q,p^*)\,.
\end{align}
Hence, for any $q(R)$, we can find $f_\theta = f^*$ for which the bound becomes tight.
\end{proof}

\subsection{Proof of \cref{prop:samplingboltz}}
\label{app:proof_samplingboltz}

\begin{mdframed}[style=MyFrame2, backgroundcolor=PastelGreenLight]
\samplingboltz*
\end{mdframed}
\begin{proof}
Recall that, for the functional $F[q]$, the functional derivative is defined through the infinitesimal change of the functional along the direction $h$ as follows
\begin{align}
    F[q + h] = F[q] + \int dx\; h(x) \underbrace{\frac{\delta F[q]}{\delta q}(x)}_{\text{derivative}} + o(\norm{h})\,,
\end{align}
which can be found as
\begin{align}
    \deriv{F[q + \eps h]}{\eps}\bigg|_{\eps=0} = \int dx\; h(x) \frac{\delta F[q]}{\delta q}(x)\,.
\end{align}

Now, consider the functional
\begin{align}
    \Phi[q,f_\theta] = \mean_{q(R)} \left[\log q(R) + E(R,f_\theta(R))\right]\,, \;\text{ subject to }\; \int dR\; q(R) = 1\,.
\end{align}
Its Lagrangian is
\begin{align}
    \mathcal{L}[q] = \mean_{q(R)} \log q(R) + \mean_{q(R)} E(R,f_\theta(R)) + \lambda \left(\int dR\;q(R) - 1\right)\,,
\end{align}
and its functional derivative is
\begin{align}
    \deriv{\mathcal{L}[q + \eps h]}{\eps}\bigg|_{\eps = 0} =~& \deriv{}{\eps}\int dR\; \left(q(R) + \eps h(R)\right) \left[\log \left(q(R) + \eps h(R)\right) + E(R,f_\theta(R)) + \lambda\right]\bigg|_{\eps = 0}\nonumber\\
    =~& \int dR\; h(R) \underbrace{\left[\log q(R) + E(R,f_\theta(R)) + \lambda\right]}_{\delta \mathcal{L} / \delta q} + \cancelto{0}{\int dR\; q(R) \frac{h(R)}{q(R)}}\,.
\end{align}
Thus, for the optimal $q$, we have
\begin{align}
    \frac{\delta \mathcal{L}}{\delta q} =~& 0\\
    \log q(R) + E(R,f_\theta(R)) =~& -\lambda\\
    q(R) =~& \frac{1}{Z_\theta}\exp(-E(R,f_\theta(R)))\,, \;\; Z_\theta = \int dR\;\exp(-E(R,f_\theta(R)))\,,
\end{align}
where the last expression we get from the normalization condition.
\end{proof}

\subsection{Proof of \cref{prop:wgradient}}
\label{app:proof_wgradient}

\begin{mdframed}[style=MyFrame2, backgroundcolor=PastelGreenLight]
\wgradient*
\end{mdframed}
\begin{proof}
Recall that the Wasserstein gradient can be derived as the vector field $v_t(x)$ that maximizes the change of the functional $F[q_t]$
\begin{align}
    \frac{d}{dt}F[q_t] = \int dx\; \deriv{q_t(x)}{t}\frac{\delta F[q_t]}{\delta q_t}\,,
\end{align}
when the density $q_t(x)$ changes according to the continuity equation
\begin{align}
    \deriv{q_t(x)}{t} = -\inner{\nabla_x}{q_t(x)v_t(x)}\,.
\end{align}
Thus, we have
\begin{align}
    \frac{d}{dt}F[q_t] = \int dx\; q_t(x)\inner{v_t(x)}{\nabla_x\frac{\delta F[q_t]}{\delta q_t}} \implies v_t(x) = \nabla_x\frac{\delta F[q_t]}{\delta q_t}\,,
\end{align}
where $\delta F[q_t]/\delta q_t$ is the functional derivative of $F[q_t]$.

Applying this reasoning for the functional $\Phi[q_t,f_\theta]$, we first have to find the functional derivative.
That is, for the functional
\begin{align}
    \Phi[q,f_\theta] = \mean_{q(R)} \left[\log q(R) + E(R,f_\theta(R))\right]\,,
\end{align}
its functional derivative is
\begin{align}
    \deriv{\Phi[q,f_\theta][q + \eps h]}{\eps}\bigg|_{\eps = 0} =~& \deriv{}{\eps}\int dR\; \left(q(R) + \eps h(R)\right) \left[\log \left(q(R) + \eps h(R)\right) + E(R,f_\theta(R))\right]\bigg|_{\eps = 0}\nonumber\\
    =~& \int dR\; h(R) \underbrace{\left[\log q(R) + E(R,f_\theta(R))\right]}_{\delta \Phi[q,f_\theta] / \delta q} + \cancelto{0}{\int dR\; q(R) \frac{h(R)}{q(R)}}\\
    \frac{\delta\Phi[q,f_\theta]}{\delta q}=~& \log q(R) + E(R,f_\theta(R))\,.
\end{align}
Thus, the PDE corresponding to the negative gradient flow is
\begin{align}
    \deriv{q_t(R)}{t} =~& -\inner{\nabla_R}{q_t(R)\left(-\nabla_R\frac{\delta\Phi[q_t,f_\theta]}{\delta q_t}\right)}\\
    =~& -\inner{\nabla_R}{q_t(R)\left(-\nabla_R\log q(R) -\nabla_R E(R,f_\theta(R))\right)} \\
    =~& -\inner{\nabla_R}{q_t(R)\left(-\nabla_R E(R,f_\theta(R))\right)} + \Delta q(R)\,.
\end{align}
This PDE is the Fokker-Planck equation, which can be efficiently simulated using the following SDE
\begin{align}
    dR_t = -\nabla_R E(R,f_\theta(R))dt + \sqrt{2}dW_t\,,
\end{align}
where $W_t$ is the standard Wiener process.
\end{proof}

\section{Architecture Details}
\label{appendix:architecture}

We use a hybrid transformer architecture composed of an SO(3)-equivariant transformer backbone  \citep{tholke_torchmd-net_2022, hassan_et-flow_2024}, and a transformer with quantum-biased attention for coefficient prediction \citep{mathiasen_reducing_2024}. 
The backbone operates on a fully-connected molecular graph, where node features encode atomic types and 3D positions, and edge features represent pairwise distances. 
It produces atomic representations which are combined with orbital-specific embeddings and passed to the transformer-based coefficient network to predict the molecular orbital coefficients.

\paragraph{Orbital tokenization and embedding}
To incorporate orbital-specific information into the coefficient prediction network, we define a fixed mapping from atomic orbital types (e.g., \texttt{1s}, \texttt{2px}, \texttt{3dxy}) to unique integer tokens. 
For each molecule, we extract atomic orbital (AO) labels from the \texttt{pySCF}, tokenize the orbital type using this mapping, and associate each orbital with the index of its parent atom. 
This allows us to extend atom-level features to the orbital level within the coefficient network.

\paragraph{Coefficient network}
The coefficient network operates on fused atomic and orbital representations, where the atomic embeddings are indexed by orbital indices to ensure that each orbital inherits the representation from its associated atom.  
The orbital tokens $T_\text{orb}$ are then embedded using a learned embedding layer, and the resulting embeddings are added to the atomic representations:
\begin{align}
    \mathbf{x}_{\text{orb}} = \mathbf{x}_{\text{atom}}[\text{index}] + \text{Embed}(T_\text{orb}).
\end{align}
The fused representations are processed by a stack of multi-head attention blocks that incorporate quantum-based biases, including the core Hamiltonian matrix \(H_{\text{core}}\) (comprising kinetic and nuclear terms), the Coulomb--exchange difference \((J - K/2)\), and the term \(S(R)H_{\text{core}}S(R)^\top\), following \citep{mathiasen_reducing_2024}.  
The final term is the transformation of the core Hamiltonian into an orthonormal basis, where \(S(R) = U\Lambda^{-\frac{1}{2}}U^\top\) is obtained via symmetric orthonormalization of the overlap matrix. 

The final output of the transformer is projected and used to construct the un-orthonormalized coefficient matrix 
\begin{equation}
    Z = \frac{QK^\top}{\sqrt{d_K}}
\end{equation}
where \(Q\) and \(K\) are the query and key matrices, and \(d_K\) is the dimensionality of the keys. 
The orbital coefficients \(C\) are then obtained by orthonormalizing \(Z\) using the reparameterization described in \cref{eq:reparam}. 
These coefficients are used to compute the total energy \(E(R, C)\) for the given configuration \(R\) with Hamiltonian \(H\), and the model is trained by minimizing this energy with respect to the network parameters \(\theta\). 
This design enables the network to distinguish among different orbitals on the same atom, and to generalize across molecules by learning shared embeddings for common orbital types.
Model hyperparameters are provided in \cref{tab:model-hparams}.

\section{Experimental Settings}
\label{appendix:exp_details}

\subsection{Dataset Details}

\cref{tab:md17-details} summarizes the relevant statistics and choice of basis set for each molecular system in the MD17 dataset.

\begin{table}[h]
\scriptsize
    \centering
    \footnotesize
    \setlength{\tabcolsep}{3pt}
    \caption{Statistics for MD17 dataset and experimental setting choices.}
    \begin{tabular}{l ccccccc}
    \toprule
        Molecule & Train ($100\%$) & Validation & Test & $\#$ Atoms & $\#$ Orbitals & Basis Set \\
        \midrule
        Ethanol         & 25,000 & 500 & 4500 & 9   & 39 & 6-31G  \\
        Malondialdehyde & 25,000 & 500 & 1478 & 9   & 53 & 6-31G  \\
        Uracil          & 25,000 & 500 & 4500 & 12  & 41 & STO-3G \\
        \bottomrule
    \end{tabular}
    \label{tab:md17-details}
\end{table}

\subsection{Training Details and Hyperparameters}
\label{appendix:train_details}

\paragraph{Training Configuration}
For all molecular systems, we explore the various configurations detailed in \cref{tab:train-details}.
We use the AdamW Optimizer with a weight decay of $10^{-6}$, and apply cosine annealing to the learning rate, decreasing it from a maximum of $3 \cdot 10^{-4}$ to a minimum $10^{-6}$ over the course of training.
Model checkpoints are selected based on the lowest validation energy computed on MD17 conformations for each molecular system.

\paragraph{Replay Buffer}
The replay buffer stores previously generated samples, which are uniformly sampled during the training of the ES model. To prevent overfitting, we maintain a buffer size of up to $2048$ samples. This buffer size also promotes samples diversity, as the stored samples are used to initialize Langevin dynamics. Once the buffer reaches capacity, it operates in a first-in-first-out (FIFO) manner, discarding the oldest samples as new ones are added.

\begin{table}[h]
\scriptsize
    \centering
    \footnotesize
    \setlength{\tabcolsep}{10pt}
    \caption{Hyperparameters for the transformer-based electronic structure model.}
    \begin{tabular}{l c}
    \toprule
        Hyper-parameter                     & Setting \\
        \midrule
        \texttt{num\_layers}                & $6$  \\
        \texttt{num\_coef\_layers}          & $6$  \\
        \texttt{hidden\_channels}           & $256$  \\ 
        \texttt{num\_heads}                 & $8$    \\
        \texttt{neighbor\_embedding}        & True   \\
        \texttt{node\_attr\_dim}            & $1$    \\
        \texttt{edge\_attr\_dim}            & $1$    \\
        \texttt{reduce\_op}                 & True  \\
        \texttt{activation}                 & SiLU  \\
        \texttt{attn\_activation}           & SiLU  \\
        \texttt{skip\_connection}           & True \\
        \midrule
        \texttt{\# param}                   & 8.4M \\
        \bottomrule
    \end{tabular}
    \label{tab:model-hparams}
\end{table}

\begin{table}[h]
\scriptsize
    \centering
    \footnotesize
    \setlength{\tabcolsep}{10pt}
    \caption{Experimental settings for training the electronic state model.}
    \begin{tabular}{llccc}
    \toprule
        Setting  & Dataset Size & Batch Size & $\#$ Iterations  & $\#$ Pretrain Iterations \\
        \midrule
        $\cD_{100\%}$                               & 25,000 & 8  & 200K & 0    \\
        $\cD_{0.1\%}$ \hspace{0.1 cm}  + SR         & 25     & 8  & 200K & 10K   \\
        $\cD_{1\%}$   \hspace{0.30 cm} + SR         & 250    & 8  & 200K & 10K   \\
        $\cD_{10\%}$  \hspace{0.18 cm} + SR         & 2,500  & 8  & 200K & 10K   \\
        $\cD_{100\%}$ \hspace{0.05 cm} + SR         & 25,000 & 8  & 200K & 10K  \\
        \bottomrule
    \end{tabular}
    \label{tab:train-details}
\end{table}

\subsection{Resources}\label{app:resources}
For all settings, we run our experiments on a cluster using one Nvidia H100 GPU for the baseline experiments, and two H100 GPUs for the self-refining method where one GPU is used to train the electronic state model, and the other GPU is used to sample from the energy of the model. 

\subsection{Runtime Analysis}\label{app:runtime-analysis}
The runtime analysis consists of three main components: (1) the time required to perform molecular dynamics (MD) simulations, (2) the time to label conformations with DFT-computed energies, and (3) the time to train the model.

To estimate the runtime for MD simulations, we follow the settings provided in \citep{schutt_unifying_2019}. 
We run simulations for 5ps to measure the wall-clock time and extrapolate to a total simulation length of 200ps. 
This estimate accounts only for integration time and does not include the cost of computing energies and forces at each step, which may vary depending on the force field or simulator used.

For DFT labeling, we use PySCF to compute total energies for a batch of 100 conformations and extrapolate the runtime to a full dataset of 25,000 conformations.

Model training time is evaluated under two scenarios. In the first, we perform supervised training using QHNet on the full pre-collected dataset. 
The total runtime includes both the time to generate the complete dataset (MD + DFT labeling) and the time to train the model. 
In the second scenario, we use self-refinement, starting from only \(10\%\) of the pre-collected dataset. 
The total runtime in this case includes the time to generate the initial \(10\%\) of conformations and the training time for self-refinement. 
Importantly, this setting eliminates the need for costly DFT energy labels, as the model is optimized directly using energy evaluations rather than by regressing to ground-truth labels.

In both scenarios, training runtimes are measured using an NVIDIA A100 GPU. A comparison of the runtimes across these settings is presented in \cref{fig:runtime-perf}.

\end{document}